\documentclass[10pt,twocolumn,letterpaper]{article}

\usepackage[pagenumbers]{cvpr} 
\usepackage{url}
\usepackage{graphicx}
\usepackage{booktabs}
\usepackage{graphicx}
\usepackage{multirow}
\usepackage{makecell}
\usepackage{colortbl}
\usepackage{xcolor}
\usepackage{amsmath}
\usepackage{amssymb}
\usepackage{mathtools}
\usepackage{enumitem}
\usepackage{microtype}
\usepackage{algorithm}
\usepackage{algorithmic}

\usepackage[pagebackref,breaklinks,colorlinks]{hyperref}

\usepackage[capitalize]{cleveref}
\crefname{section}{Sec.}{Secs.}
\Crefname{section}{Section}{Sections}
\Crefname{table}{Table}{Tables}
\crefname{table}{Tab.}{Tabs.}

\def\cvprPaperID{*****} 
\def\confName{CVPR}
\def\confYear{2022}

\begin{document}

\title{Zoomed In, Diffused Out: Towards Local Degradation-Aware Multi-Diffusion for Extreme Image Super-Resolution}

\author{Brian B. Moser$^{1, 2}$, Stanislav Frolov$^{1, 2}$, Tobias C. Nauen$^{1, 2}$, Federico Raue$^{1}$, Andreas Dengel$^{1, 2}$\\
$^{1}$German Research Center for Artificial Intelligence \\
$^{2}$University of Kaiserslautern-Landau\\
{\tt\small first.last@dfki.de}
}
\maketitle

\begin{abstract}
   Large-scale, pre-trained Text-to-Image (T2I) diffusion models have gained significant popularity in image generation tasks and have shown unexpected potential in image Super-Resolution (SR). 
   However, most existing T2I diffusion models are trained with a resolution limit of 512$\times$512, making scaling beyond this resolution an unresolved but necessary challenge for image SR.
   In this work, we introduce a novel approach that, for the first time, enables these models to generate 2K, 4K, and even 8K images without any additional training. 
   Our method leverages MultiDiffusion, which distributes the generation across multiple diffusion paths to ensure global coherence at larger scales, and local degradation-aware prompt extraction, which guides the T2I model to reconstruct fine local structures according to its low-resolution input. 
   These innovations unlock higher resolutions, allowing T2I diffusion models to be applied to image SR tasks without limitation on resolution.
\end{abstract}

\section{Introduction}
Image Super-Resolution (SR) is vital for a wide range of real-world applications, including satellite imaging, medical diagnostics, and consumer photography, where high-resolution outputs (\textit{e.g.}, 2K, 4K, or 8K) are essential for capturing fine details and ensuring clarity \cite{zhan2021achieving, shi2016real, schreiber2017audiences, adami2016social, boudraa2020improving}. 
Although SR methods, particularly those using local operations like CNNs, have made significant progress, handling complex degradation in Low-Resolution (LR) inputs remains a persistent challenge \cite{moser2023hitchhiker, liu2022blind, moser2024diffusion}.

Recently, diffusion models, particularly pre-trained Text-to-Image (T2I) diffusion models, have revolutionized image generation tasks \cite{frolov2021adversarial, lugmayr2022repaint, zhang2023adding, wu2023uncovering}. 
Originally designed for creative applications like text-guided image synthesis, these models have demonstrated strong potential in image SR, especially in handling 4x scaling and beyond, where hallucinating fine details becomes essential \cite{moser2024waving, moser2024diffusion}. 
Diffusion-based SR models, such as SR3, DiffBIR, and SRDiff, have already achieved impressive results by generating realistic details at these scales using conventional SR datasets and training them from scratch \cite{saharia2022image, li2022srdiff, lin2024diffbir}.

\begin{figure}[t!]
    \begin{center}
        \includegraphics[width=\columnwidth]{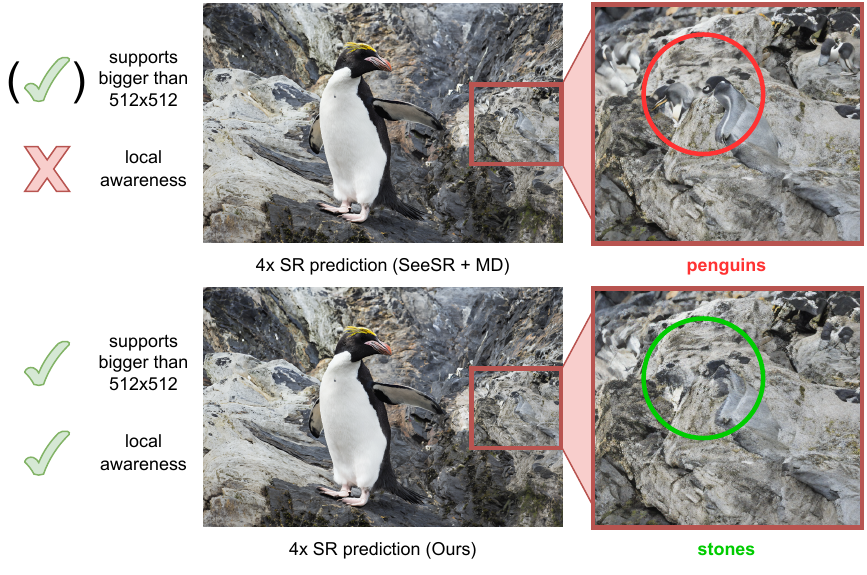}
        
        \caption{\label{fig:idea}
        Comparison of 4x Super-Resolution (SR) predictions using SeeSR + MultiDiffusion (MD) and our proposed method. While MD unlocks higher resolutions beyond 512$\times$512, our proposed strategy of extracting local degradation-aware prompts ensures local detail awareness, improving fine-grained structure restoration, as demonstrated in the stones regions.
        }
    \end{center} 
\end{figure}

On the other side, the development of T2I models accelerates, as exemplified by astonishing applications like Dall-E and StableDiffusion \cite{rombach2022high, betker2023improving}.
Driven by training on increasingly large and diverse datasets, they present a great resource for image SR \cite{ganguli2022predictability, kaplan2020scaling}. 
By repurposing their vast generative capabilities, these models can unlock new possibilities for High-Resolution (HR) image enhancement.
Inspired by this idea, methods like StableSR, PASD, and SeeSR have repurposed pre-trained T2I models to SR, achieving impressive results in limited-resolution scenarios \cite{wang2023exploiting, yang2023pixel, wu2023seesr}.

Yet, all T2I diffusion models repurposed for image SR are limited by a resolution of 512$\times$512, which is impractical for real-world applications and a major limitation \cite{frolov2024spotdiffusion, du2024demofusion, bar2023multidiffusion, quattrini2024merging}.

In this work, we introduce a novel solution that, for the first time, enables pre-trained T2I diffusion models to generate SR images at resolutions of 2K, 4K, and even 8K, which we coin extreme image SR, without any additional training. 
Our approach leverages two core components: (1) MultiDiffusion, which distributes the image generation process across multiple diffusion paths to ensure global coherence, and (2) local degradation-aware prompt extraction, which provides fine-grained control over local structures, allowing the model to effectively restore detailed textures and semantic consistency.
By integrating local degradation-aware prompt extraction, our method ensures consistent semantic accuracy across all image scales, setting a new standard for extreme image SR.

In summary, our key contributions are as follows:
\begin{itemize}
    \item We introduce MultiDiffusion to image SR, which enables scaling T2I diffusion models (trained for 512x512) beyond 512x512 by coordinating multiple diffusion paths for global coherence.
    \item We propose local degradation-aware prompt extraction, which enhances the ability to preserve local details and structural integrity.
    \item We demonstrate, for the first time, the capability of pre-trained T2I models to generate 2K, 4K, and 8K images without training and set an initial benchmark for extreme image SR alongside classical methods.
\end{itemize}

\section{Related Work}

This section briefly reviews the current field of image SR and how T2I is exploited for image SR. 

\subsection{Image SR}
Image SR has seen significant advancements through the development of CNNs. 
Classical CNN-based SR models (including vision transformers), such as RRDB, ESRGAN, SwinIR, and HAT, excel in upscaling images to any resolution due to their local operations \cite{wang2018esrgan, liang2021swinir, chen2023activating}. 
They leverage local receptive fields, enabling them to process and train on relatively small image patches (\textit{e.g.}, 192x192) while generalizing well to larger images. 
This property makes them highly scalable for SR tasks across various resolutions.

More recently, diffusion models have emerged as powerful alternatives for SR, particularly in handling higher scaling factors such as 4x and beyond. 
Models like SR3, DiffBIR, and SRDiff have demonstrated strong capabilities in hallucinating fine details required at these scales \cite{saharia2022image, lin2024diffbir, li2022srdiff}. 
However, these diffusion-based SR models typically need to be trained on the target size and are often limited to resolutions of 512x512 due to architectural and training constraints, which restrict their flexibility and practicality compared to CNN-based methods that can be applied to any resolution.

In contrast, our approach introduces a novel method to scale diffusion models beyond 512x512 without retraining, which we termed extreme image SR (\textit{i.e.}, 2K, 4K, and 8K).

\subsection{Exploiting T2I for SR}
Text-to-Image (T2I) conditioning has emerged as a promising approach for enhancing Super-Resolution (SR) tasks by leveraging the capabilities of pre-trained T2I models. 
Fine-tuning these models, with the addition of specialized encoders suited for SR, allows for integrating textual descriptions into image enhancement. 
This fusion of textual information can provide a richer guidance source, potentially improving the accuracy and contextual relevance of synthesized images in SR applications.

For instance, \textit{Wang et al.} introduced StableSR, which exploits text guidance by incorporating a time-aware encoder trained alongside a frozen Latent Diffusion Model (LDM) \cite{wang2023exploiting, rombach2022high}. 
This framework utilizes trainable spatial feature transform layers to condition the model based on input images. 
An optional controllable feature wrapping module further enhances StableSR's adaptability, allowing for fine-grained user control. 
The design of this module draws inspiration from CodeFormer, contributing to StableSR’s flexibility in addressing diverse user needs and preferences \cite{zhou2022towards}.

Similarly, \textit{Yang et al.} proposed Pixel-Aware Stable Diffusion (PASD), which advances the conditioning process by embedding text descriptions of LR inputs using a CLIP text encoder \cite{yang2023pixel, radford2021learning}. 
This method improves the model’s capacity to synthesize images with greater precision by embedding contextual information from textual sources, enhancing the overall fidelity and relevance of the generated images.

Concurrently, other methods like SeeSR explore similar T2I conditioning frameworks, while XPSR takes this concept further by merging different levels of semantic text encodings \cite{wu2023seesr, qu2024xpsr}. 
XPSR combines high-level encodings (image content) with low-level encodings (quality perception, sharpness, noise, and other LR image distortions), further refining the SR results through a more nuanced understanding of both content and perceptual quality.

\section{Methodology}
\begin{figure*}[t!]
    \begin{center}
        \includegraphics[width=\textwidth]{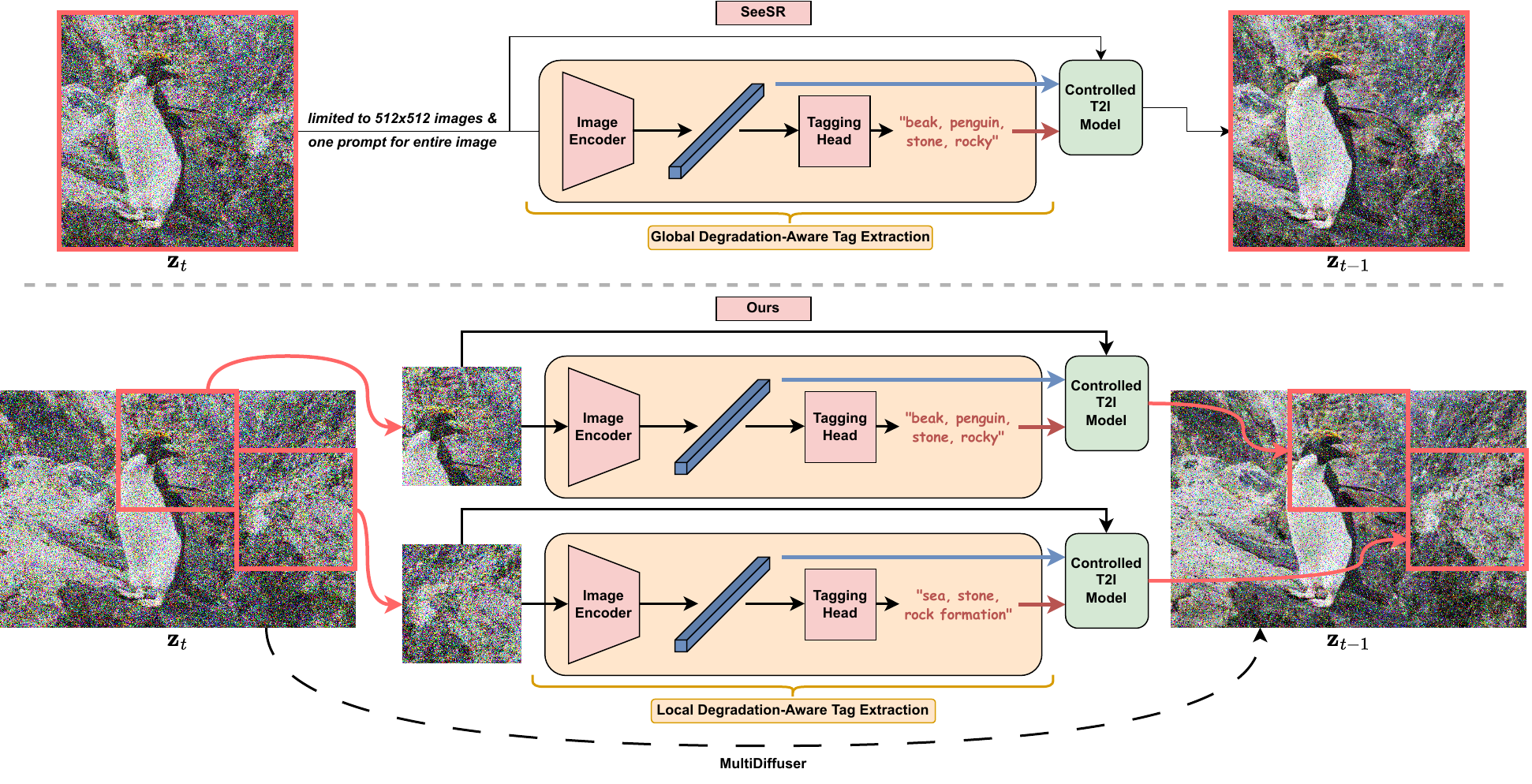}
        \caption{\label{fig:main}
        Illustration of SeeSR \textbf{(top)} compared to our local degradation-aware method \textbf{(bottom)}. 
        While SeeSR is limited to a fixed image size of 512$\times$512, our method can technically upscale to any resolution due to two components: MultiDiffusion (MD) and local degradation-aware prompt extraction.
        The MD process is applied to overlapping tiles. 
        Without local degradation-aware prompt extraction, the classifier guidance generates hallucinated details based on global prompts that describe the entire image, leading to inconsistencies in local tile content. 
        Our approach incorporates local tag extraction and, thereby, provides tile-specific prompts, ensuring more accurate and coherent detail generation across the entire image.
        }
    \end{center} 
\end{figure*}

This section introduces our novel approach to extreme image SR by leveraging pre-trained T2I diffusion models. 
For the first time, our method enables T2I models repurposed for SR to achieve resolutions of 2K, 4K, and 8K without additional training but with local coherence. 
The approach is built upon two core innovations: MultiDiffusion, which ensures global coherence at high resolutions, and local degradation-aware prompt extraction, which enhances local detail recovery. 
\autoref{fig:main} illustrates the overall concept of our method.

\subsection{MultiDiffusion Process}

MultiDiffusion (MD) allows the model to generate high-resolution images by distributing the image synthesis across multiple diffusion paths \cite{bar2023multidiffusion, frolov2024spotdiffusion, quattrini2024merging}. 
At each stage, the latent feature map is divided into overlapping tiles, each of which undergoes a separate diffusion process. 
This technique maintains global coherence by sharing overlapping information between adjacent tiles. 
The final image is synthesized by merging the outputs from these multiple paths, ensuring consistency in both global structure and local detail. 
Unlike traditional T2I diffusion methods, which are limited to 512$\times$512 pixels, this process enables resolutions of 2K and beyond without any retraining or fine-tuning.

More concretely, let \(\Phi: \mathcal{L} \times \mathcal{Y} \rightarrow \mathcal{L}\) be a pre-trained T2I model that operates in the latent space \(\mathcal{L} = \mathbb{R}^{W\times H\times C} \) and condition space \(\mathcal{Y}\), \textit{e.g.}, textual prompts.
Given a noisy latent representation \(L_T \sim \mathcal{N}(\mathbf{0}, \mathbf{I})\) and a condition \(y \in \mathcal{Y}\), the T2I diffusion model \(\Phi\) produces a sequence of latents starting from \(L_T\) and gradually denoising it towards the clean latent representation \(L_0\) that can be decoded to a HR approximation:
\begin{equation}
L_T, L_{T-1}, \ldots, L_0 \quad \text{such that} \quad L_{t-1} = \Phi(L_t \mid y).
\end{equation}

Since we want to unlock the generation of images larger than 512$\times$512 pixels (\textit{i.e.}, latent codes larger than 64$\times$64), our goal is the generation latent representations $M_T, M_{T-1}, \ldots, M_0$ in a new latent space \(\mathcal{M} = \mathbb{R}^{W'\times H'\times C} \), with $W'\geq W, H'\geq H$, using the same pre-trained model \(\Phi\) without any retraining or fine-tuning. 
Traditionally, models pre-trained on fixed-size latents cannot be directly applied to produce latents of arbitrary sizes \cite{mei2024bigger, chen2024image}.

To address this, MD extends the diffusion process by applying a joint diffusion approach, where multiple overlapping latent windows are merged via averaging. 
More formally, we define $n$ mappings $F_i: \mathcal{M}\in \mathbb{R}^{W'\times H'\times C} \to \mathcal{L}\in \mathbb{R}^{W\times H\times C}$ with $i \in \{1, \ldots ,n\}$, which map (or crop) the larger latent space $\mathcal{M}$ into $n$ latent representations of the original size $W \times H$ (\textit{i.e.}, 64$\times$64).

The number of overlapping latent crops $n$ is defined as $n=\frac{W^{\prime}-W}{\omega}+1$, where $\omega$ represents the stride between adjacent cropping windows. 
With these mappings, the denoising process is applied independently to each cropped latent window. 
Subsequently, the latent representations are stitched back to the original size  \(\mathcal{M} = \mathbb{R}^{W'\times H'\times C} \), \textit{i.e.} $M_{t-1} = \text{MultiDiffuser}(\{L_{t-1}^{i}\}^n_{i=1})$, which averages the overlapping latent regions.

\subsection{Local Degradation-Aware Prompt Extraction}
As demonstrated in the introductory example in \autoref{fig:idea}, solely applying MD often results in images with excessive hallucinated details. 
This occurs because the global prompt used for classifier guidance in the diffusion process applies uniformly across all tiles. 
While the prompt effectively describes the global structure of the image, it lacks the local content specificity needed for each tile. 
Consequently, the model attempts to reconstruct fine details in every tile based on global information, leading to over-hallucination and inconsistencies in local structures.

More formally, let \(y \in \mathcal{Y}\) be a condition drawn from a prompt extractor.
As $F_i$ is guiding the diffusion model $\Phi$ with the same prompt $y$, \textit{i.e.}, $\Phi(L^i_t \mid y)$ for all possible windows $i \in \{1, \ldots, n\}$, the larger latent feature map resulting from this MD process will have the conditioning information $y$ infused at every spatial position.
Hence, global prompts alone are insufficient for reconstructing region-specific content, as they lack the granularity to capture the unique local details within each patch.

To ensure the coherence of local structure and details, we propose a local degradation-aware prompt extraction, which describes the local degradation patterns in LR inputs (see next Section for concrete extractor). 
In more detail, if $n$ is the number of maps (or crops), we propose to condition the MD process with a set of $n$ conditions, \textit{i.e.}, $\{y_1, \ldots, y_n\}$, where $F_i$ uses the condition $y_i$.
These prompts guide the diffusion model in reconstructing detailed textures and structures during the SR process. 
By leveraging these prompts, our method effectively reduces artifacts and noise, particularly at high resolutions (\textit{e.g.}, 2K, 4K, and 8K), and ensures accurate restoration of fine-grained details.
As a result, our approach, as outlined in \autoref{alg:md_with_local_prompts}, significantly mitigates the common issues of over-hallucination and visual artifacts that occur when relying solely on global prompt guidance.

\begin{algorithm}[t]
  \caption{Local Degradation-Aware MultiDiffusion}
  \label{alg:md_with_local_prompts}
  \begin{algorithmic}[1]
  \REQUIRE $\Phi$ \COMMENT{pre-trained T2I diffusion model} \\
  \hspace{0.8cm} $\varphi$ \COMMENT{pre-trained prompt extractor} \\
  \hspace{0.8cm} $\{F_i\}_{i=1}^n$ \COMMENT{non-overlapping latent mappings} \\
  \hspace{0.8cm} $\{I_i\}_{i=1}^n$ \COMMENT{corresponding image patches} \\
  \hspace{0.8cm} $M_T \sim \mathcal{N}(\mathbf{0}, \mathbf{I})$ \COMMENT{noisy initialization} \\
  \FOR{each window $i = 1, \ldots, n$}
        \STATE $y_i \leftarrow \varphi \left( I_i \right)$ \COMMENT{extract local prompts}
  \ENDFOR 
  \FOR{$t = T, T-1, \ldots, 0$}
          
      \FOR{each window $i = 1, \ldots, n$}
          \STATE $L_{t-1}^{i} \leftarrow \Phi (F_i \left(M_t\right) \mid y_i)$ \COMMENT{denoise window}
      \ENDFOR
      \STATE $M_{t-1} \leftarrow \text{MultiDiffuser}(\{L_{t-1}^{i}\}^n_{i=1})$
  \ENDFOR
  \STATE \textbf{return} $M_0$
  \end{algorithmic}
\end{algorithm}

\subsection{Exploiting Pre-Trained Prompt Extractors}
Our method is designed to be flexible and compatible with any pre-trained T2I diffusion model and prompt extraction strategy. 
Still, in this work, we specifically build upon the pre-trained models utilized by SeeSR \cite{wu2023seesr}.
Specifically, we utilize the pre-trained StableDiffusion V2.0 model, along with its 8$\times$ compression VAE, and the Degradation-Aware Prompt Extractor (DAPE), which itself is a fine-tuned tag-style prompt extraction model, \textit{i.e.}, RAM \cite{rombach2022high, esser2021taming, wu2023seesr, zhang2024recognize}. 
These components have proven effective for real-world SR tasks, allowing us to leverage their robust generative capabilities for higher-resolution image synthesis.

For the MD process, we adopt the standard configuration of dividing the latent space into 64x64 patches \cite{bar2023multidiffusion}. 
This approach ensures efficient global coherence across the image while maintaining the fine-grained structure. Inspired by SpotDiffusion, we reduce the overlap between patches by setting the stride to 32, ensuring a balance between computational efficiency and image consistency \cite{frolov2024spotdiffusion}. 
Notably, we chose not to apply the non-overlapping striding strategy proposed in SpotDiffusion, as it significantly degraded the SR results for LR inputs by failing to capture necessary contextual information across patches.

For the local degradation-aware prompt extraction, we apply DAPE on the corresponding tiles in image space (\textit{i.e.}, extracted from 512$\times$512 image patches) that would undergo the individual diffusion steps in $F_i$ in latent space. 
This extraction ensures that each patch receives locally relevant prompts, enhancing fine detail restoration. 
Interestingly, this process results in averaging overlapping latent patches generated under varying prompt conditions. 
This dynamic use of local prompts helps prevent inconsistencies between adjacent patches and avoids over-hallucination, which is common when using global prompts in MD \cite{bar2023multidiffusion}.

\section{Experimental Setup}
In this section, we detail the datasets, models, and evaluation metrics used to assess the performance of our proposed method, comparing it against both classical and diffusion-based SR models across various extreme-resolution tasks.
The code for our experiments can be found on GitHub \footnote{\url{https://github.com/Brian-Moser/zido}}, which complements the official implementation of SeeSR.

\begin{figure*}[ht!]
    \begin{center}
        \includegraphics[width=\textwidth]{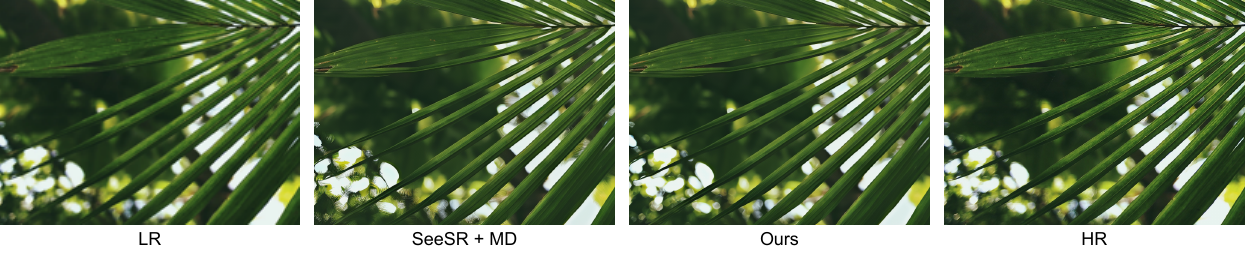}
        
        \caption{\label{fig:qual_comp_2}
        Qualitative comparison of a 2K image (899; DIV2K Val) between LR, SeeSR+MD (PSNR$\uparrow$:25.273, SSIM$\uparrow$:0.769, LPIPS$\downarrow$:0.130), our method (PSNR$\uparrow$:\textbf{26.217}, SSIM$\uparrow$:\textbf{0.794}, LPIPS$\downarrow$:\textbf{0.103}), and HR.
        In general, we observe that our method reconstructs details in background objects better than SeeSR+MD (see light patterns in the lower left corner).
        }
    \end{center} 
\end{figure*}

\subsection{Datasets}
For our experiments, we utilized the DIV2K validation set as well as the Test4K and Test8K datasets introduced by ClassSR \cite{agustsson2017ntire, kong2021classsr}. 
These datasets provide a robust benchmark for SR tasks, allowing us to evaluate the performance of our method across a range of high resolutions (\textit{i.e.}, 2K, 4K, and 8K), which we coin extreme image SR. 

Since our approach fully leverages pre-trained T2I models without any additional training, we did not use the classical SR training datasets like DIV2K (train) and Flickr2K \cite{agustsson2017ntire, timofte2018ntire}.
Our focus on enabling extreme SR at resolutions far beyond 512×512 also led us to exclude conventional SR evaluation datasets, such as Set5, Set14, BSDS100, Manga109, and Urban100 \cite{bevilacqua2012low, zeyde2010single, martin2001database, matsui2017sketch, huang2015single}. 
These datasets are commonly used in SR research but are limited in resolution, typically capping at 512×512, making them unsuitable for testing the full capabilities of our method \cite{bar2023multidiffusion, rombach2022high, wu2023seesr, yang2023pixel}. 
By focusing on extreme resolutions, we aim to demonstrate the scalability and versatility of our approach in supporting any-resolution SR with pre-trained models, particularly for 2K, 4K, and 8K.

For the scaling factor, we selected a 4$\times$ magnification, meaning that the HR images are four times larger in spatial dimensions than their LR counterparts.
This scaling factor is common in diffusion-based SR benchmarks and is generally considered challenging \cite{moser2024diffusion, moser2023hitchhiker}.

\subsection{Models \& Metrics}
We compare a variety of standard SR approaches, including regression-based methods like EDSR, RRDB, and CAR, GAN-based methods such as RankSRGAN and ESRGAN, as well as the normalizing flow approach SRFlow \cite{lugmayr2020srflow,ma2020structure,lim2017enhanced,soh2019natural,ledig2017photo,zhang2019ranksrgan,wang2018recovering,wang2018esrgan}. Additionally, we include diffusion models that are explicitly trained for image SR, such as SR3+YODA, SRDiff, and DiWa, to provide a broader comparison \cite{moser2024waving, saharia2022image, li2022srdiff, moser2023yoda}.

To evaluate the quality of the reconstructed images, we use both pixel-based metrics, namely PSNR and SSIM, as well as the perceptual quality metric LPIPS \cite{moser2023hitchhiker, moser2024diffusion}. Pixel-based metrics measure structural similarity and reconstruction accuracy, while LPIPS provides a more human-perception-aligned evaluation of image quality, allowing us to assess both fidelity and perceptual realism.
Note that classical SR approaches typically exhibit much higher pixel-based scores than diffusion models, as diffusion models are prone to hallucinated details \cite{saharia2022image}. 

Additionally, lower scores are expected in our method since the mentioned models are explicitly SR-trained, whereas we repurpose pre-trained T2I models without SR-training (thereby modifying SeeSR with MD as baseline).

\section{Results}
This section presents qualitative and quantitative comparisons between our approach and SeeSR modified with MD (leading in SR-repurposed T2I methods). 
Since no pre-trained T2I models have been previously tested for extreme image SR (resolutions larger than 512$\times$512), SeeSR with MD serves as our primary baseline for evaluation.

\begin{figure}[ht!]
    \begin{center}
        \includegraphics[width=\columnwidth]{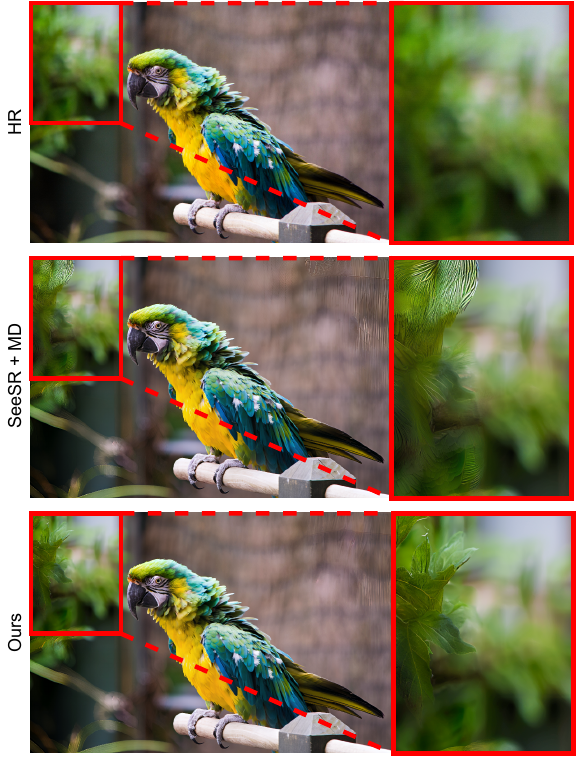}
        
        \caption{\label{fig:qual_comp}
        Qualitative comparison (886; DIV2K Val) between
        SeeSR+MD (PSNR$\uparrow$:26.252, SSIM$\uparrow$:0.766, LPIPS$\downarrow$:0.123) and our approach (PSNR$\uparrow$:\textbf{28.115}, SSIM$\uparrow$:\textbf{0.802}, LPIPS$\downarrow$:\textbf{0.091}).
        The global tags were ``balustrade, bird, blue, fence, green, macaw, parrot, perch, pole, rail, sit, stand, yellow''.
        While SeeSR+MD hallucinates bird patterns on the leaves in the background due to global prompt guidance, our approach preserves local coherence by reconstructing leaves more naturally. 
        However, although the background is content-wise accurate, our method introduces more fine-grained, blurry-free details than those present in the original HR image. 
        }
    \end{center} 
\end{figure}

\subsection{Qualitative Results}

To assess the effectiveness of our proposed approach visually, we focus on challenging yet visible cases from the DIV2K validation set to highlight the improvements in fine detail preservation, local coherence, and overall perceptual quality.
As shown in \autoref{fig:qual_comp_2}, our method demonstrates significant improvements in recovering fine details and generating more coherent structures compared to SeeSR+MD. 
In the 2K resolution image, our approach not only preserves the global structure but also reconstructs background objects, such as the light patterns in the lower-left corner, with more accuracy and clarity (\textit{i.e.}, without artifacts). 
This is due to the local degradation-aware prompt extraction, which enables more precise local detail reconstruction, avoiding the hallucination of irrelevant global features.

\begin{figure}[ht!]
    \begin{center}
        \includegraphics[width=\columnwidth]{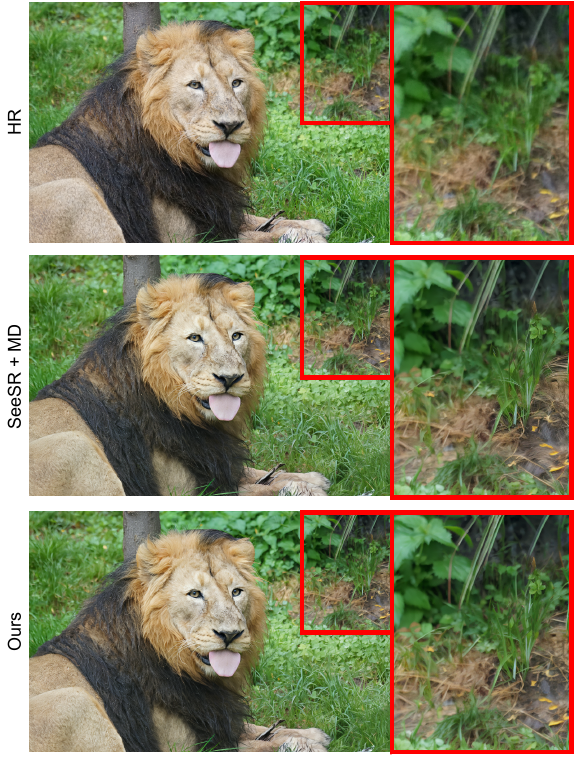}
        
        \caption{\label{fig:qual_comp_3}
        Qualitative comparison (809; DIV2K Val) between
        SeeSR+MD (PSNR$\uparrow$:25.107, SSIM$\uparrow$:0.598, LPIPS$\downarrow$:0.076) and our approach (PSNR$\uparrow$:\textbf{25.627}, SSIM$\uparrow$:\textbf{0.621}, LPIPS$\downarrow$:\textbf{0.069}).
        The global tags were ``animal, break, floor, grass, green, lay, lion, lush, man, mane, mouth, relax, tree''.
        Similarly, SeeSR+MD hallucinates fur-like patterns in the brown dirt, leading to artifacts that degrade the visual quality and contribute to its inferior performance compared to our approach, which preserves the natural texture of the dirt more effectively.
        Once again, our method generates finer, sharper details that surpass the level found in the HR image. 
        }
    \end{center} 
\end{figure}

In \autoref{fig:qual_comp} and \autoref{fig:qual_comp_3}, we further illustrate how our method handles the presence of background textures by taking a closer look at zoomed-in regions. 
In the first example (\autoref{fig:qual_comp}), SeeSR+MD hallucinates feather-like patterns in the leaves due to the global prompt guidance, leading to inconsistent results (see wavy structures above the leaf). 
By contrast, our method maintains local coherence and reconstructs the leaves more naturally and accurately. 
For the second example (\autoref{fig:qual_comp_3}), we can observe fur-like patterns in the dirt in SeeSR, contrary to ours.
Although both methods occasionally introduce more fine-grained details than those in the original HR image, our method avoids artifacts and hallucinations, ensuring perceptually consistent results.

Overall, our qualitative results demonstrate that the local degradation-aware prompt extraction significantly outperforms SeeSR+MD in generating high-quality super-resolved images, particularly at extreme resolutions. 
The improvements are especially notable in preserving local textures and avoiding over-hallucination, which is a common issue in MD-based models guided by global prompts.
The addition of local degradation-aware prompts not only ensures more consistent textures but also provides a sharper, artifact-free reconstruction of fine details, making our approach particularly advantageous for complex scenes where global prompts alone fail to maintain coherence.

\subsection{Quantitative Results}
To thoroughly evaluate the performance of our method, we conducted experiments on the DIV2K validation set, as well as the Test4K and Test8K datasets. 
Our results are compared against various existing SR methods, including regression-based, GAN-based, and diffusion-based approaches (trained explicitly on SR data).

As shown in \autoref{tab:div2k_results}, our method outperforms SeeSR equipped with MD regarding perceptual quality metrics such as LPIPS while achieving competitive results on pixel-based metrics like PSNR and SSIM. 
Specifically, for 4x SR on DIV2K, our method achieved a PSNR of 24.34, SSIM of 0.68, and LPIPS of 0.108, representing a clear improvement in image quality compared to SeeSR+MD.

We further evaluated our method on the challenging Test4K and Test8K datasets to demonstrate its scalability to extreme resolutions, a crucial requirement for modern SR tasks. 
As shown in \autoref{tab:test4k_results}, our method consistently outperformed SeeSR+MD on the 4x SR task for the Test4K dataset, achieving a PSNR of 26.871, SSIM of 0.735, and LPIPS of 0.0920. 
Similarly, on the more demanding Test8K dataset (see \autoref{tab:test8k_results}), our method achieved a PSNR of 23.505, SSIM of 0.653, and LPIPS of 0.0796, clearly surpassing SeeSR+MD across both pixel-based and perceptual quality metrics.

In contrast to SeeSR, our approach not only achieves superior perceptual results but also excels in producing sharper, artifact-free images with consistently higher PSNR and SSIM values. 
By scaling beyond the typical 512x512 limit, our method contributes significantly towards making T2I practical for real-world SR applications.

\subsection{User Study}
Inspired by \textit{Saharia et al.} (SR3 \cite{saharia2022image}), we conducted a 2-
alternative forced-choice user study.
We asked 25 subjects ``Which of the two images is a better high-quality version of the low-resolution
image in the middle?''
We selected 36 random 2K test images from DIV2K Val (12 for each method: RRDB, SeeSR+MD, and ours), which were center-cropped to $512\times512$ for better visual examination.

\begin{figure}[t!]
    \begin{center}
        \includegraphics[width=\columnwidth]{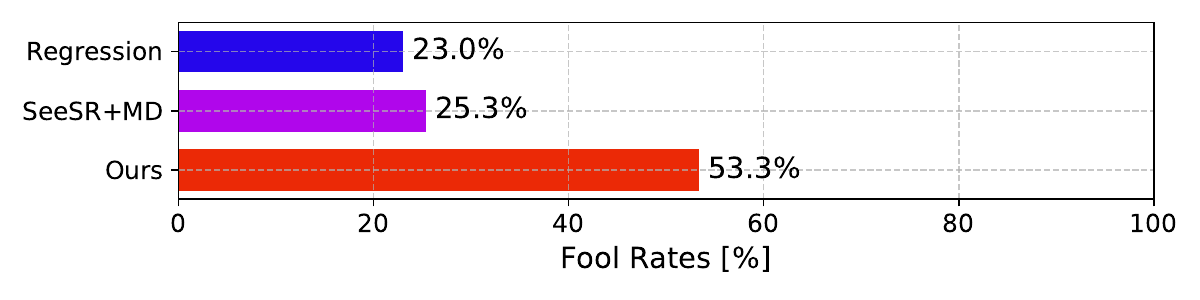}
        \caption{\label{fig:user_study}DIV2K Val fool rates (higher is better; photo-realistic samples yield a fool rate of 50\%). Regression (RRDB), SeeSR+MD, and our outputs are compared against ground truth. 
        }
    \end{center} 
\end{figure}

The result of our user study is shown in \autoref{fig:user_study}.
Our method not only surpasses RRDB and SeeSR+MD, achieving more than double their fool rates, but it also attains a fool rate close to 50\%, rendering its outputs nearly indistinguishable from HR ground-truth images.

\begin{table}
\center
\small
\resizebox{\columnwidth}{!}{%
\begin{tabular}{c l c c c }
\toprule
\textbf{Type} & \textbf{Methods} & \textbf{PSNR} $\uparrow$ & \textbf{SSIM} $\uparrow$& \textbf{LPIPS} $\downarrow$ \\ 
\midrule
\multirow{4}{*}{Regression} & Bicubic & 26.70 & 0.77 & 0.409  \\
& EDSR & 28.98 & 0.83 & 0.270  \\
& RRDB & 29.44 & 0.84 & 0.253  \\ 
& CAR & 32.82 & 0.88 & -  \\ 
\midrule
\multirow{2}{*}{GANs} & RankSRGAN & 26.55 & 0.75 & 0.128  \\ 
& ESRGAN & 26.22 & 0.75 & 0.124  \\
\midrule
NFs & SRFlow & 27.09 & 0.76 & 0.120  \\
\midrule
Diffusion & SR3 + YODA & 27.24 & 0.77 & 0.127 \\
(SR-trained) & SRDiff & 27.41 & 0.79 & 0.136 \\ 
& DiWa  & 28.09 &  0.78 & 0.104 \\
\midrule
Diffusion & SeeSR + MD & 24.28 & 0.67 & 0.110 \\  
(T2I-trained)& \textbf{Ours} & \textbf{24.34} & \textbf{0.68} & \textbf{0.108} \\
\bottomrule
\end{tabular}
}
\caption{Quantitative results for 4x SR on the DIV2K validation set. Our method outperforms SeeSR+MD in both perceptual quality (LPIPS) and pixel-based metrics (PSNR and SSIM). Regression-, Normalizing Flow- and GAN-based SR models are also included for comparison, alongside diffusion models trained specifically for SR tasks (DIV2K+Flickr2K), contrary to SeeSR and our method.
}
\label{tab:div2k_results}
\end{table}

\begin{table}[t]
\center
\resizebox{\columnwidth}{!}{%
\begin{tabular}{c l c c c }
\toprule
\textbf{Type} & \textbf{Methods}& \textbf{PSNR} $\uparrow$ & \textbf{SSIM} $\uparrow$& \textbf{LPIPS} $\downarrow$ \\ 
\midrule
\multirow{2}{*}{Regression} & Bicubic & 31.749 & 0.843 & 0.0817 \\
& RRDB & 33.931 & 0.887 & 0.0536 \\
\midrule
GANs & ESRGAN & 30.609 & 0.810 & 0.0348 \\
\midrule
Diffusion & SeeSR + MD & 26.808 & 0.727 & 0.0972  \\
(T2I-trained)& \textbf{Ours} & \textbf{26.871} & \textbf{0.735} & \textbf{0.0920}\\ 
\bottomrule
\end{tabular}
}
\caption{Results of 4x SR on the Test4K dataset. Our method consistently outperforms SeeSR+MD, achieving higher PSNR, SSIM, and better perceptual quality (LPIPS), showcasing its strength in handling extreme-resolution SR tasks.
}
\label{tab:test4k_results}
\end{table}

\begin{table}[t]
\center
\resizebox{\columnwidth}{!}{%
\begin{tabular}{c l c c c }
\toprule
\textbf{Type} & \textbf{Methods}& \textbf{PSNR} $\uparrow$ & \textbf{SSIM} $\uparrow$& \textbf{LPIPS} $\downarrow$ \\ 
\midrule
\multirow{2}{*}{Regression} & Bicubic & 26.103 & 0.750 & 0.1264 \\
& RRDB & 28.012 & 0.817 & 0.0754 \\
\midrule
GANs & ESRGAN & 24.789 & 0.723 & 0.0462 \\
\midrule
Diffusion & SeeSR + MD & 23.473 & 0.647 & 0.0822  \\
(T2I-trained)& \textbf{Ours} & \textbf{23.505} & \textbf{0.653} & \textbf{0.0796}\\ 
\bottomrule
\end{tabular}
}
\caption{Results of 4x SR on the Test8K dataset. Our method surpasses SeeSR+MD across all metrics - PSNR, SSIM, and LPIPS - demonstrating its scalability and superior performance at high resolutions.
}
\label{tab:test8k_results}
\end{table}

\subsection{Prompt Analysis}
We analyze the textual information added by our local degradation-aware prompt extraction in DIV2K Val by counting unique tags.
For our method and the same LR image, we count common tags between the MD paths as one. 
Thus, a higher count implies more tailored guidance for different image areas, supporting our hypothesis that local prompts contribute to better restoration of complex structures.

\begin{figure}[t!]
    \begin{center}
        \includegraphics[width=\columnwidth]{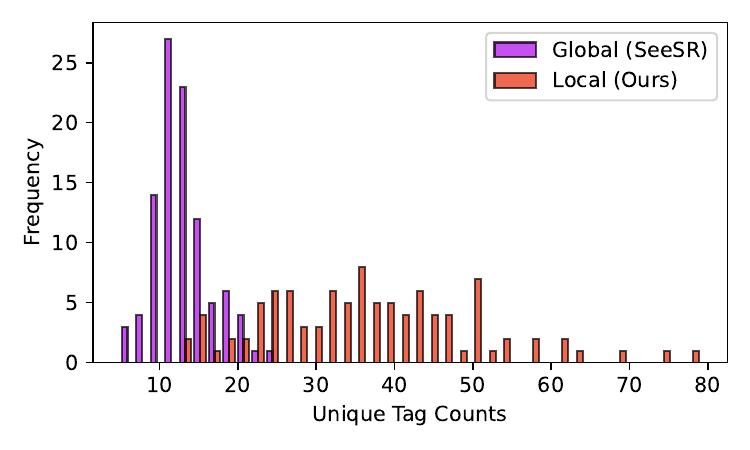}
        
        \caption{\label{fig:tag_counts}
        The distribution of unique tag counts for DIV2K Val. 
        Tags appearing across multiple image patches are counted as one. 
        Our method generates more unique tags overall, with the highest frequency peaking at 35 compared to SeeSR, which peaks around 10.
        The global tag extraction spans around 5-20 unique tags, while we produce 10-80.
        }
    \end{center} 
\end{figure}

\begin{figure}[t!]
    \begin{center}
        \includegraphics[width=.99\columnwidth]{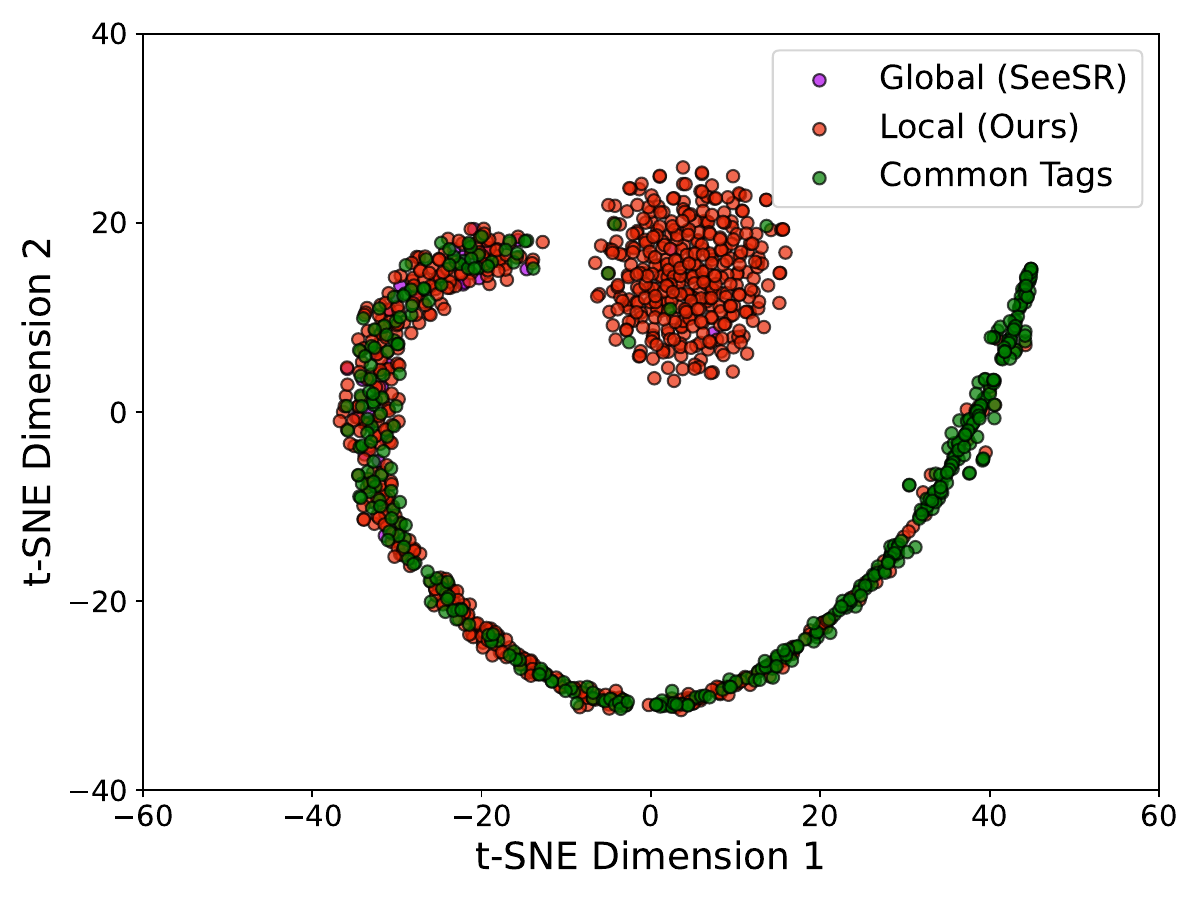}
        \caption{\label{fig:clusters}t-SNE of Word2Vec embedded tags by our method and SeeSR, alongside common unique tags. Our local tag extraction includes similar tags (line) and adds more unique tags (circle).
        }
    \end{center} 
\end{figure}

As shown in \autoref{fig:tag_counts}, our proposed local degradation-aware prompt extraction method generates significantly more unique tags than SeeSR.
Also, it highlights that we generate a wider range of tags, allowing for a more precise description. 
Naturally, increased tag diversity improves local detail description across the image and is further evidenced in \autoref{fig:clusters}: Our local degradation-aware prompt extraction includes the majority of tags produced by SeeSR but also adds distinct ones, as underscored by the red cluster.

This enhancement in local coherence is one of the primary reasons our approach outperforms SeeSR in quantitative metrics (such as LPIPS) and qualitative visual comparisons, as detailed in the previous results.

\section{Limitation \& Future Work}
While our proposed local degradation-aware prompt extraction method significantly improves the performance of T2I diffusion models for any-resolution image SR, it still has room for improvement compared to traditional SR methods, which are explicitly designed and trained for SR. 

Future work could explore a hybrid approach that combines the strengths of traditional SR models with T2I models to address the performance difference compared to SR-trained methods like RRDB and ESRGAN. 
For instance, using a GAN or CNN-based model to generate an initial coarse super-resolved image and refinement using a diffusion model could lead to even better quality.

\section{Conclusion}
This work introduced a novel approach for extreme image SR by leveraging pre-trained T2I diffusion models. 
Through our proposed MD process and local degradation-aware prompt extraction, we enabled the generation of high-quality images at resolutions of 2K, 4K, and 8K without any additional training. 
By distributing the generation process across multiple diffusion paths, our method ensures global coherence, while local prompt extraction enhances fine-grained detail restoration. 
This addresses the limitations of traditional T2I models exploited for image SR, making them applicable to a broad range of tasks where clarity and precision at large scales are critical.

Our extensive experiments show that our method significantly outperforms the baseline SeeSR+MD in both qualitative and quantitative metrics, particularly in challenging high-resolution scenarios. 
Enabling any high-resolutions, maintaining local coherence, and avoiding over-hallucination while generating visually accurate details sets a new benchmark for high-resolution SR using pre-trained T2I models.

\section*{Acknowledgements}
This work was supported by the BMBF projects SustainML (Grant 101070408), Albatross (Grant 01IW24002) and by Carl Zeiss Foundation through the Sustainable Embedded AI project (P2021-02-009).

{\small
\bibliographystyle{ieee_fullname}
\bibliography{egbib}

\begin{thebibliography}{10}\itemsep=-1pt

\bibitem{adami2016social}
Elisabetta Adami and Carey Jewitt.
\newblock Social media and the visual, 2016.

\bibitem{agustsson2017ntire}
Eirikur Agustsson and Radu Timofte.
\newblock Ntire 2017 challenge on single image super-resolution: Dataset and study.
\newblock In {\em CVPRW}, pages 126--135, 2017.

\bibitem{bar2023multidiffusion}
Omer Bar-Tal, Lior Yariv, Yaron Lipman, and Tali Dekel.
\newblock Multidiffusion: Fusing diffusion paths for controlled image generation.
\newblock 2023.

\bibitem{betker2023improving}
James Betker, Gabriel Goh, Li Jing, Tim Brooks, Jianfeng Wang, Linjie Li, Long Ouyang, Juntang Zhuang, Joyce Lee, Yufei Guo, et~al.
\newblock Improving image generation with better captions.
\newblock {\em Computer Science. https://cdn. openai. com/papers/dall-e-3. pdf}, 2(3):8, 2023.

\bibitem{bevilacqua2012low}
Marco Bevilacqua, Aline Roumy, Christine Guillemot, and Marie~Line Alberi-Morel.
\newblock Low-complexity single-image super-resolution based on nonnegative neighbor embedding.
\newblock 2012.

\bibitem{boudraa2020improving}
Sawsen Boudraa, Ahlem Melouah, and Hayet~Farida Merouani.
\newblock Improving mass discrimination in mammogram-cad system using texture information and super-resolution reconstruction.
\newblock {\em Evolving Systems}, 11(4):697--706, 2020.

\bibitem{chen2023activating}
Xiangyu Chen, Xintao Wang, Jiantao Zhou, Yu Qiao, and Chao Dong.
\newblock Activating more pixels in image super-resolution transformer.
\newblock In {\em CVPR}, pages 22367--22377, 2023.

\bibitem{chen2024image}
Yinbo Chen, Oliver Wang, Richard Zhang, Eli Shechtman, Xiaolong Wang, and Michael Gharbi.
\newblock Image neural field diffusion models.
\newblock In {\em CVPR}, pages 8007--8017, 2024.

\bibitem{du2024demofusion}
Ruoyi Du, Dongliang Chang, Timothy Hospedales, Yi-Zhe Song, and Zhanyu Ma.
\newblock Demofusion: Democratising high-resolution image generation with no \$\$\$.
\newblock In {\em CVPR}, pages 6159--6168, 2024.

\bibitem{esser2021taming}
Patrick Esser, Robin Rombach, and Bjorn Ommer.
\newblock Taming transformers for high-resolution image synthesis.
\newblock In {\em CVPR}, pages 12873--12883, 2021.

\bibitem{frolov2021adversarial}
Stanislav Frolov, Tobias Hinz, Federico Raue, J{\"o}rn Hees, and Andreas Dengel.
\newblock Adversarial text-to-image synthesis: A review.
\newblock {\em Neural Networks}, 144:187--209, 2021.

\bibitem{frolov2024spotdiffusion}
Stanislav Frolov, Brian~B Moser, and Andreas Dengel.
\newblock Spotdiffusion: A fast approach for seamless panorama generation over time.
\newblock {\em arXiv preprint arXiv:2407.15507}, 2024.

\bibitem{ganguli2022predictability}
Deep Ganguli, Danny Hernandez, Liane Lovitt, Amanda Askell, Yuntao Bai, Anna Chen, Tom Conerly, Nova Dassarma, Dawn Drain, Nelson Elhage, et~al.
\newblock Predictability and surprise in large generative models.
\newblock In {\em 2022 ACM Conference on Fairness, Accountability, and Transparency}, 2022.

\bibitem{huang2015single}
Jia-Bin Huang, Abhishek Singh, and Narendra Ahuja.
\newblock Single image super-resolution from transformed self-exemplars.
\newblock In {\em CVPR}, pages 5197--5206, 2015.

\bibitem{kaplan2020scaling}
Jared Kaplan, Sam McCandlish, Tom Henighan, Tom~B Brown, Benjamin Chess, Rewon Child, Scott Gray, Alec Radford, Jeffrey Wu, and Dario Amodei.
\newblock Scaling laws for neural language models.
\newblock {\em arXiv preprint arXiv:2001.08361}, 2020.

\bibitem{kong2021classsr}
Xiangtao Kong, Hengyuan Zhao, Yu Qiao, and Chao Dong.
\newblock Classsr: A general framework to accelerate super-resolution networks by data characteristic.
\newblock In {\em CVPR}, pages 12016--12025, 2021.

\bibitem{ledig2017photo}
Christian Ledig, Lucas Theis, Ferenc Husz{\'a}r, Jose Caballero, Andrew Cunningham, Alejandro Acosta, Andrew Aitken, Alykhan Tejani, Johannes Totz, Zehan Wang, et~al.
\newblock Photo-realistic single image super-resolution using a generative adversarial network.
\newblock In {\em CVPR}, 2017.

\bibitem{li2022srdiff}
Haoying Li, Yifan Yang, Meng Chang, Shiqi Chen, Huajun Feng, Zhihai Xu, Qi Li, and Yueting Chen.
\newblock Srdiff: Single image super-resolution with diffusion probabilistic models.
\newblock In {\em Neurocomputing}, 2022.

\bibitem{liang2021swinir}
Jingyun Liang, Jiezhang Cao, Guolei Sun, Kai Zhang, Luc Van~Gool, and Radu Timofte.
\newblock Swinir: Image restoration using swin transformer.
\newblock In {\em ICCV}, pages 1833--1844, 2021.

\bibitem{lim2017enhanced}
Bee Lim, Sanghyun Son, Heewon Kim, Seungjun Nah, and Kyoung Mu~Lee.
\newblock Enhanced deep residual networks for single image super-resolution.
\newblock In {\em CVPRW}, pages 136--144, 2017.

\bibitem{lin2024diffbir}
Xinqi Lin, Jingwen He, Ziyan Chen, Zhaoyang Lyu, Ben Fei, Bo Dai, Wanli Ouyang, Yu Qiao, and Chao Dong.
\newblock Diffbir: Towards blind image restoration with generative diffusion prior.
\newblock {\em arXiv preprint arXiv:2308.15070}, 2023.

\bibitem{liu2022blind}
Anran Liu, Yihao Liu, Jinjin Gu, Yu Qiao, and Chao Dong.
\newblock Blind image super-resolution: A survey and beyond.
\newblock {\em IEEE TPAMI}, 2022.

\bibitem{lugmayr2022repaint}
Andreas Lugmayr, Martin Danelljan, Andres Romero, Fisher Yu, Radu Timofte, and Luc Van~Gool.
\newblock Repaint: Inpainting using denoising diffusion probabilistic models.
\newblock In {\em CVPR}, 2022.

\bibitem{lugmayr2020srflow}
Andreas Lugmayr, Martin Danelljan, Luc Van~Gool, and Radu Timofte.
\newblock Srflow: Learning the super-resolution space with normalizing flow.
\newblock In {\em ECCV}, 2020.

\bibitem{ma2020structure}
Cheng Ma, Yongming Rao, Yean Cheng, Ce Chen, Jiwen Lu, and Jie Zhou.
\newblock Structure-preserving super resolution with gradient guidance.
\newblock In {\em CVPR}, 2020.

\bibitem{martin2001database}
David Martin, Charless Fowlkes, Doron Tal, and Jitendra Malik.
\newblock A database of human segmented natural images and its application to evaluating segmentation algorithms and measuring ecological statistics.
\newblock In {\em ICCV}, volume~2, pages 416--423. IEEE, 2001.

\bibitem{matsui2017sketch}
Yusuke Matsui, Kota Ito, Yuji Aramaki, Azuma Fujimoto, Toru Ogawa, Toshihiko Yamasaki, and Kiyoharu Aizawa.
\newblock Sketch-based manga retrieval using manga109 dataset.
\newblock {\em Multimedia tools and applications}, 76:21811--21838, 2017.

\bibitem{mei2024bigger}
Kangfu Mei, Zhengzhong Tu, Mauricio Delbracio, Hossein Talebi, Vishal~M Patel, and Peyman Milanfar.
\newblock Bigger is not always better: Scaling properties of latent diffusion models.
\newblock {\em arXiv preprint arXiv:2404.01367}, 2024.

\bibitem{moser2023yoda}
Brian~B Moser, Stanislav Frolov, Federico Raue, Sebastian Palacio, and Andreas Dengel.
\newblock Yoda: You only diffuse areas. an area-masked diffusion approach for image super-resolution.
\newblock {\em arXiv preprint arXiv:2308.07977}, 2023.

\bibitem{moser2024waving}
Brian~B Moser, Stanislav Frolov, Federico Raue, Sebastian Palacio, and Andreas Dengel.
\newblock Waving goodbye to low-res: A diffusion-wavelet approach for image super-resolution.
\newblock In {\em 2024 International Joint Conference on Neural Networks (IJCNN)}, pages 1--8. IEEE, 2024.

\bibitem{moser2023hitchhiker}
Brian~B Moser, Federico Raue, Stanislav Frolov, Sebastian Palacio, J{\"o}rn Hees, and Andreas Dengel.
\newblock Hitchhiker's guide to super-resolution: Introduction and recent advances.
\newblock {\em IEEE TPAMI}, 45(8):9862--9882, 2023.

\bibitem{moser2024diffusion}
Brian~B Moser, Arundhati~S Shanbhag, Federico Raue, Stanislav Frolov, Sebastian Palacio, and Andreas Dengel.
\newblock Diffusion models, image super-resolution, and everything: A survey.
\newblock {\em IEEE Transactions on Neural Networks and Learning Systems}, 2024.

\bibitem{qu2024xpsr}
Yunpeng Qu, Kun Yuan, Kai Zhao, Qizhi Xie, Jinhua Hao, Ming Sun, and Chao Zhou.
\newblock Xpsr: Cross-modal priors for diffusion-based image super-resolution.
\newblock {\em arXiv:2403.05049}, 2024.

\bibitem{quattrini2024merging}
Fabio Quattrini, Vittorio Pippi, Silvia Cascianelli, and Rita Cucchiara.
\newblock Merging and splitting diffusion paths for semantically coherent panoramas.
\newblock {\em arXiv preprint arXiv:2408.15660}, 2024.

\bibitem{radford2021learning}
Alec Radford, Jong~Wook Kim, Chris Hallacy, Aditya Ramesh, Gabriel Goh, Sandhini Agarwal, Girish Sastry, Amanda Askell, Pamela Mishkin, Jack Clark, et~al.
\newblock Learning transferable visual models from natural language supervision.
\newblock pages 8748--8763. PMLR, 2021.

\bibitem{rombach2022high}
Robin Rombach, Andreas Blattmann, Dominik Lorenz, Patrick Esser, and Bj{\"o}rn Ommer.
\newblock High-resolution image synthesis with latent diffusion models.
\newblock In {\em CVPR}, 2022.

\bibitem{saharia2022image}
Chitwan Saharia, Jonathan Ho, William Chan, Tim Salimans, David~J Fleet, and Mohammad Norouzi.
\newblock Image super-resolution via iterative refinement.
\newblock {\em IEEE TPAMI}, 2022.

\bibitem{schreiber2017audiences}
Maria Schreiber.
\newblock Audiences, aesthetics and affordances analysing practices of visual communication on social media.
\newblock {\em Digital Culture \& Society}, 3(2):143--164, 2017.

\bibitem{shi2016real}
Wenzhe Shi, Jose Caballero, Ferenc Husz{\'a}r, Johannes Totz, Andrew~P Aitken, Rob Bishop, Daniel Rueckert, and Zehan Wang.
\newblock Real-time single image and video super-resolution using an efficient sub-pixel convolutional neural network.
\newblock In {\em CVPR}, pages 1874--1883, 2016.

\bibitem{soh2019natural}
Jae~Woong Soh, Gu~Yong Park, Junho Jo, and Nam~Ik Cho.
\newblock Natural and realistic single image super-resolution with explicit natural manifold discrimination.
\newblock In {\em CVPR}, 2019.

\bibitem{timofte2018ntire}
Radu Timofte, Shuhang Gu, Jiqing Wu, and Luc Van~Gool.
\newblock Ntire 2018 challenge on single image super-resolution: Methods and results.
\newblock In {\em CVPRW}, 2018.

\bibitem{wang2023exploiting}
Jianyi Wang, Zongsheng Yue, Shangchen Zhou, Kelvin~CK Chan, and Chen~Change Loy.
\newblock Exploiting diffusion prior for real-world image super-resolution.
\newblock {\em arXiv preprint}, 2023.

\bibitem{wang2018recovering}
Xintao Wang, Ke Yu, Chao Dong, and Chen~Change Loy.
\newblock Recovering realistic texture in image super-resolution by deep spatial feature transform.
\newblock In {\em CVPR}, pages 606--615, 2018.

\bibitem{wang2018esrgan}
Xintao Wang, Ke Yu, Shixiang Wu, Jinjin Gu, Yihao Liu, Chao Dong, Yu Qiao, and Chen Change~Loy.
\newblock Esrgan: Enhanced super-resolution generative adversarial networks.
\newblock In {\em ECCV}, 2018.

\bibitem{wu2023uncovering}
Qiucheng Wu, Yujian Liu, Handong Zhao, Ajinkya Kale, Trung Bui, Tong Yu, Zhe Lin, Yang Zhang, and Shiyu Chang.
\newblock Uncovering the disentanglement capability in text-to-image diffusion models.
\newblock In {\em CVPR}, pages 1900--1910, 2023.

\bibitem{wu2023seesr}
Rongyuan Wu, Tao Yang, Lingchen Sun, Zhengqiang Zhang, Shuai Li, and Lei Zhang.
\newblock Seesr: Towards semantics-aware real-world image super-resolution.
\newblock {\em arXiv:2311.16518}, 2023.

\bibitem{yang2023pixel}
Tao Yang, Rongyuan Wu, Peiran Ren, Xuansong Xie, and Lei Zhang.
\newblock Pixel-aware stable diffusion for realistic image super-resolution and personalized stylization.
\newblock {\em arXiv preprint arXiv:2308.14469}, 2023.

\bibitem{zeyde2010single}
Roman Zeyde, Michael Elad, and Matan Protter.
\newblock On single image scale-up using sparse-representations.
\newblock In {\em International conference on curves and surfaces}, pages 711--730. Springer, 2010.

\bibitem{zhan2021achieving}
Zheng Zhan, Yifan Gong, Pu Zhao, Geng Yuan, Wei Niu, Yushu Wu, Tianyun Zhang, Malith Jayaweera, David Kaeli, Bin Ren, et~al.
\newblock Achieving on-mobile real-time super-resolution with neural architecture and pruning search.
\newblock In {\em ICCV}, pages 4821--4831, 2021.

\bibitem{zhang2023adding}
Lvmin Zhang, Anyi Rao, and Maneesh Agrawala.
\newblock Adding conditional control to text-to-image diffusion models.
\newblock In {\em ICCV}, pages 3836--3847, 2023.

\bibitem{zhang2019ranksrgan}
Wenlong Zhang, Yihao Liu, Chao Dong, and Yu Qiao.
\newblock Ranksrgan: Generative adversarial networks with ranker for image super-resolution.
\newblock In {\em ICCV}, 2019.

\bibitem{zhang2024recognize}
Youcai Zhang, Xinyu Huang, Jinyu Ma, Zhaoyang Li, Zhaochuan Luo, Yanchun Xie, Yuzhuo Qin, Tong Luo, Yaqian Li, Shilong Liu, et~al.
\newblock Recognize anything: A strong image tagging model.
\newblock In {\em CVPR}, pages 1724--1732, 2024.

\bibitem{zhou2022towards}
Shangchen Zhou, Kelvin Chan, Chongyi Li, and Chen~Change Loy.
\newblock Towards robust blind face restoration with codebook lookup transformer.
\newblock {\em NeurIPS}, 35, 2022.

\end{thebibliography}
}

\end{document}